\documentclass{bmvc2k}
\usepackage{soul}
\usepackage{wrapfig, blindtext}
\usepackage[symbol]{footmisc}
\usepackage{eurosym}

\title{Human-Machine Collaborative Design for Accelerated Design of Compact Deep Neural Networks for Autonomous Driving}

\addauthor{Mohammad Javad Shafiee}{}{1}
\addauthor{Mirko Nentwig,Yohannes Kassahun}{}{2}
\addauthor{Francis Li, Stanislav Bochkarev, \mbox{Akif Kamal}, \mbox{David Dolson}, \mbox{Secil~Altintas}, \mbox{Arif Virani}}{}{3}
\addauthor{ Alexander Wong}{}{1}

\addinstitution{  Waterloo AI Institute, University of Waterloo \& DarwinAI Corp., Waterloo, Ontario, Canada.}
\addinstitution{ Audi Electronics Venture GmbH (AEV), Gaimersheim, Germany.}
\addinstitution{ DarwinAI Corp., Waterloo, Ontario, Canada.}

\runninghead{}{BMVC Workshop on Visual AI and Entrepreneurship (VAIE)}

\begin{document}
\maketitle
\vspace{-0.1in}
\begin{abstract}
An effective deep learning development process is critical for widespread industrial adoption, particularly in the automotive sector. A typical industrial deep learning development cycle involves customizing and re-designing an off-the-shelf network architecture to meet the operational requirements of the target application, leading to considerable trial and error work by a machine learning practitioner. This approach greatly impedes development with a long turnaround time and the unsatisfactory quality of the created models. As a result, a development platform that can aid engineers in greatly accelerating the design and production of compact, optimized deep neural networks is highly desirable.  In this joint industrial case study, we study the efficacy of the GenSynth AI-assisted AI design platform for accelerating the design of custom, optimized deep neural networks for autonomous driving through human-machine collaborative design. We perform a quantitative examination by evaluating 10 different compact deep neural networks produced by GenSynth for the purpose of object detection via a NASNet-based user network prototype design, targeted at a low-cost GPU-based accelerated embedded system.  Furthermore, we quantitatively assess the talent hours and GPU processing hours used by the GenSynth process and three other approaches based on the typical industrial development process. In addition, we quantify the annual cloud cost savings for comprehensive testing using networks produced by GenSynth.  Finally, we assess the usability and merits of the GenSynth process through user feedback.  The findings of this case study showed that GenSynth is easy to use and can be effective at accelerating the design and production of compact, customized deep neural network.
\end{abstract}

%-------------------------------------------------------------------------
\section{Introduction}
\vspace{-0.1in}
\label{sec:intro}
   
\begin{figure}
\centering
\includegraphics[width=0.9\linewidth]{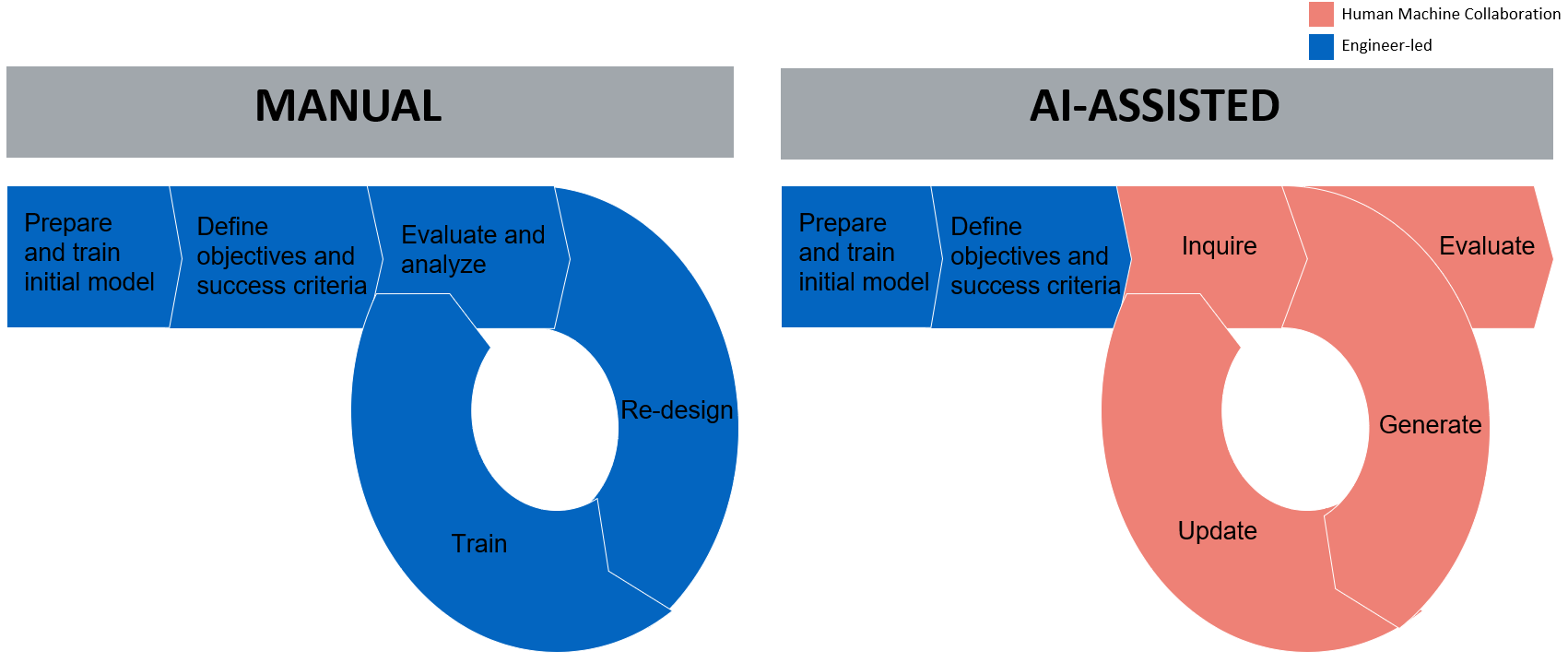}
\caption{\footnotesize (left) Typical deep learning development cycle in industry vs. (right) an AI-assisted deep learning development cycle using the GenSynth platform}
\vspace{-0.2in}
\label{fig:cycle}
\end{figure}

Recent successes in deep learning~\cite{amodei2016deep,szegedy2016rethinking,redmon2017yolo9000} have led to considerable interest in widespread adoption across industrial sectors, particularly the automotive sector. As such, the establishment of an effective deep learning development process becomes critical.  A typical industrial deep learning development cycle is shown in Figure~\ref{fig:cycle}(left), where an off-the-shelf network architecture is customized, re-designed, and evaluated in an iterative fashion to create a model that meets the needs of a specific application. There are several critical challenges with this approach that greatly impede both turnaround time and the quality of the created models. First, it is very difficult to choose the appropriate off-the-shelf model for a tailored custom task, and thus the off-the-shelf model used as the initial prototype might not be the best model architecture for the particular task at hand since it was not designed for that specific task in the first place. As a result, such a model often does not provide optimal performance for the task.  Second,  off-the-shelf models are typically designed to be as generic as possible (e.g., 1000 different generic classes). Therefore, such models typically have very high architectural and computational complexity, making them impractical to deploy on edge devices, such as those used for autonomous driving, which have limited computational, memory, and energy resources. 

Due to the aforementioned reasons, the typical industrial development cycle of customizing off-the-shelf models to meet the operational requirements of a specific task becomes extremely time-consuming and, in many cases, impossible to achieve as it requires machine learning engineers to devote a significant amount of time to modifying the network architecture for their specific task. One of the main goals is to design high-performance deep learning models that are tailored for embedded automotive environments where one must consider not only the task and dataset, but also the memory, computational, energy,  and hardware constraints in the design process.  Having a development platform that can aid engineers in accelerating the design and production of compact, optimized deep neural networks that meet or exceed operational requirements is highly desirable for an automotive company that must constantly design new and improved deep neural networks to power their autonomous driving initiatives.  

In this joint industrial case study between AEV and DarwinAI, we explore the efficacy of the GenSynth AI-assisted AI design platform for accelerating the design of custom, optimized deep neural networks for autonomous driving through human-machine collaborative design.  By combining the ingenuity and experience of a human designer with the meticulousness and raw speed of AI, the potential of such a design platform can not only produce better and more compact networks tailored for specific tasks more quickly, but also help designers gain better insights and improve their own ability to design deep neural networks, leading the an improved AI-assisted deep learning development cycle (see Figure~\ref{fig:cycle}(right)). 

\begin{figure}
\centering
\includegraphics[width=0.4\linewidth]{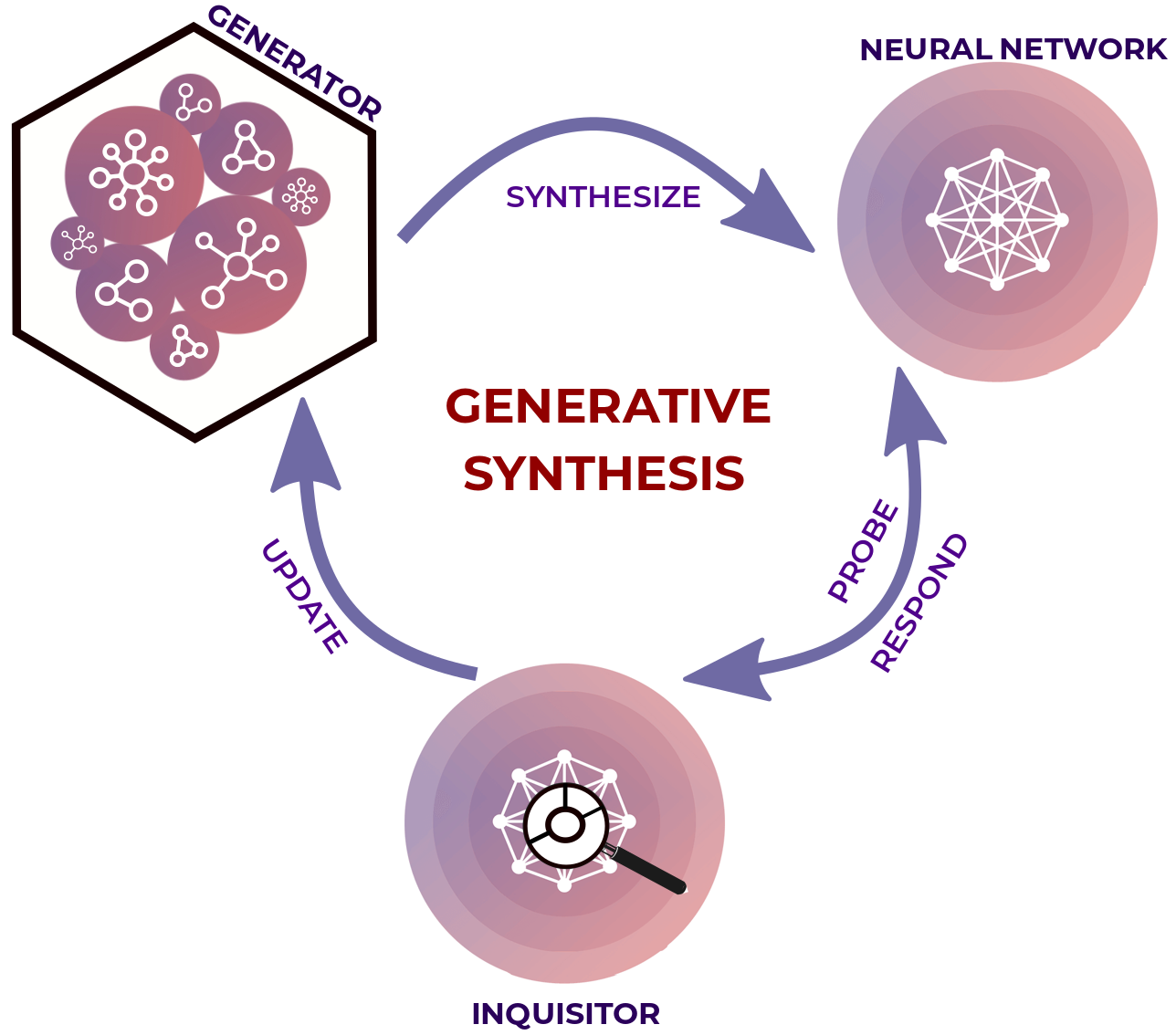} \\
\caption{\footnotesize Overview of the Generative Synthesis concept powering the GenSynth AI-assisted AI platform~\cite{wong2018ferminets}.}
\vspace{-0.2in}
\label{fig:gensynth}
\end{figure}

\vspace{-0.1in}
\section{Generative Synthesis}
\vspace{-0.1in}
The concept of Generative Synthesis, which powers the GenSynth platform, is designed around human-machine collaborative design. This achieves a  very fine-grain macro-architecture and micro-architecture design exploration via machine-driven design exploration while learning from a human-specified network design prototype along with data and design requirements.  The overarching goal of Generative Synthesis~\cite{wong2018ferminets} is to learn a generator $\mathcal{G}$ that, given a set of seeds $S$, can generate networks $\left\{N_s|s \in S\right\}$ that maximize a universal performance function $\mathcal{U}$ (e.g.,~\cite{Wong2018_Netscore}) while satisfying quantitative human-specified requirements defined via an indicator function $1_r(\cdot)$.  This can be formulated as a constrained optimization problem,
\vspace{-0.1in}
\begin{equation}
\mathcal{G}  = \max_{\mathcal{G}}~\mathcal{U}(\mathcal{G}(s))~~\textrm{subject~to}~~1_r(\mathcal{G}(s))=1,~~\forall s \in S.
\label{optimization}
\vspace{-0.1in}
\end{equation}

Given the intractability of Eq.~\ref{optimization}, an approximate solution $\mathcal{G}$ to the constrained optimization problem posed is achieved by leveraging the interplay between a generator-inquisitor pair that works in tandem to obtain improved insights about deep neural networks as well as learn to generate highly efficient networks in a cyclical manner.  As shown in Figure~\ref{fig:gensynth}, a generator is first learned based on a user network design prototype, data, and requirements (size, accuracy tolerance, hardware-level requirements like channel multiplicity and data precision, etc.) and is used to generate deep neural networks.  An inquisitor probes the generated deep neural networks and the corresponding reactionary response is observed.  Based on these observations, the inquisitor learns at a foundational level about the architectural efficiencies of the generated deep neural networks and updates the generator based on the insights it gains.  This leads to an improved generator that is then used to generate new deep neural networks. The aforementioned process of generating, probing, observing, and updating is repeated over cycles, resulting in a sequence of improving generators.  We take full advantage of this interesting phenomenon by leveraging this sequence of generators to synthesize different highly-efficient deep neural networks that satisfy these requirements but with different trade-offs between modeling accuracy and efficiency.

\vspace{-0.1in}
\section{Case Study Analysis }
\vspace{-0.1in}
In this case study, the goal is to accelerate the design of a compact deep convolutional neural network for object detection in autonomous driving using GenSynth.  The user design prototype specified by AEV is based on a NASNet Faster-RCNN network design, which possesses a highly complex architecture consisting of 352 convolutional layers with a focus on detecting 10 different types of automotive-relevant classes. The 10 classes are as follow: Buses, Cars, Trucks, Motorcycles, Pedestrians, Traffic, Signs, Trains, Fire Hydrants, Bicycles, Traffic Lights. The target hardware are   GPU-based systems.

\begin{figure}
    \centering
    \includegraphics[width=9cm]{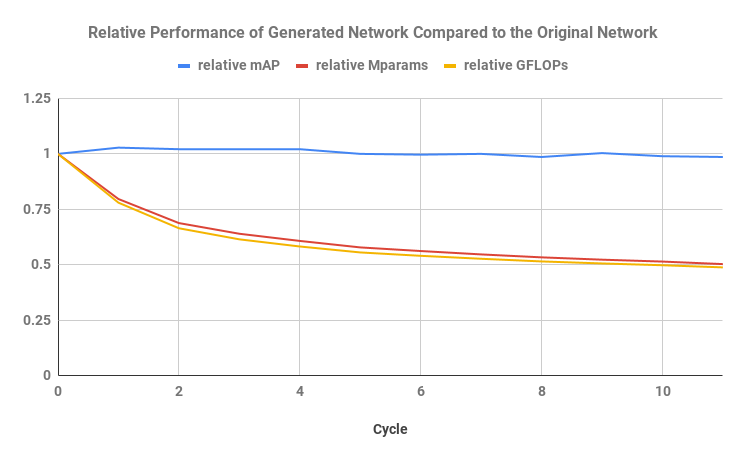}
     \vspace{-0.5cm}
    \caption{ \footnotesize Performance comparison for the original user design prototype and a generated network architecture by the GenSynth platform.}
    \label{fig:nasnetfaster_rcnn_result}
\end{figure}

For this particular study, GenSynth learns and generates 10 different compact deep neural networks based on the user design prototype and data.  It can be clearly observed in Figure~\ref{fig:nasnetfaster_rcnn_result} that the deep neural networks generated by GenSynth achieved either a higher or similar mean average precision (mAP) as the user design prototype, while being significantly more compact as well as achieving significantly lower computational costs. In particular, the most compact generated deep neural network possesses half the number of parameters and float-point operations than the user-designed prototype while providing the same accuracy level as the original user design prototype.  The experimental results were conducted on a single GPU machine; however, the processing time would have been lower if a multi-GPU configuration was used.
Table~\ref{tab:nasnet_result}  shows the analytic metrics for the original user design prototype and one of the generated networks produced by the GenSynth platform. The reported running-time inference is measured based on an Nvidia V100 GPU as well as an embedded GPU device in this study.  As seen, the GenSynth platform can generate new network architectures that are noticeably more compact and have a lower computational cost, leading to significant speed gains on both the Nvidia V100 GPU as well as the embedded GPU device.    Figure~\ref{fig:nasnetfaster_rcnn_result} demonstrates the relative performance of the network architectures generated by the GenSynth platform compared to the original user design prototype.

\begin{table}[]
    \centering
    \caption{\footnotesize Performance comparison for the original user design prototype and the generated network architecture by the GenSynth platform running on two different processing units. }
    \setlength\tabcolsep{3pt} 
    \begin{tabular}{|c||c|c|c|c|}
     \hline
         \bf Model &\bf Hardware &  \bf GFLops &\bf Mparams &\bf Relative Speed-up\\  \hline
         \bf Original Network& --- & 683.6& 105.4 & 1$\times$ \\    
         \bf  GenSynth-Generated &  Embedded GPU& 352.6& 55.2 & 1.47$\times$ \\     
         \bf  GenSynth-Generated & V100 GPU & 352.6& 55.2 & 1.8$\times$ \\
        \hline
    \end{tabular}
    
    \label{tab:nasnet_result}
\end{table}

To further illustrate the benefits of GenSynth for helping AEV accelerate the design of a custom compact deep neural network through human-machine collaborative design, let us contrast two different approaches as shown in Figure~\ref{fig:cycle}:

\begin{enumerate}
\item \textbf{The conventional manual process.} In this approach, the machine learning practitioner needs to evaluate the current model and then, based on the evaluation, re-design the model, including changing the number of layers or tuning the number of filters in each layer and, in the last step, train the new network to gain full modeling accuracy. This process is applied iteratively until the final model satisfies the requirements.

\item \textbf{The GenSynth process.} In  this approach, after defining the requirements, the network model is passed to the GenSynth process which inquires from the current model. This generates a new network architecture and automatically updates in a loop until the criteria for success are met. However, the practitioner can be involved in the process whenever it is necessary. 
\end{enumerate}
 
 \begin{figure}
\begin{tabular}{cc}
\multicolumn{2}{c}{\includegraphics[width=10cm]{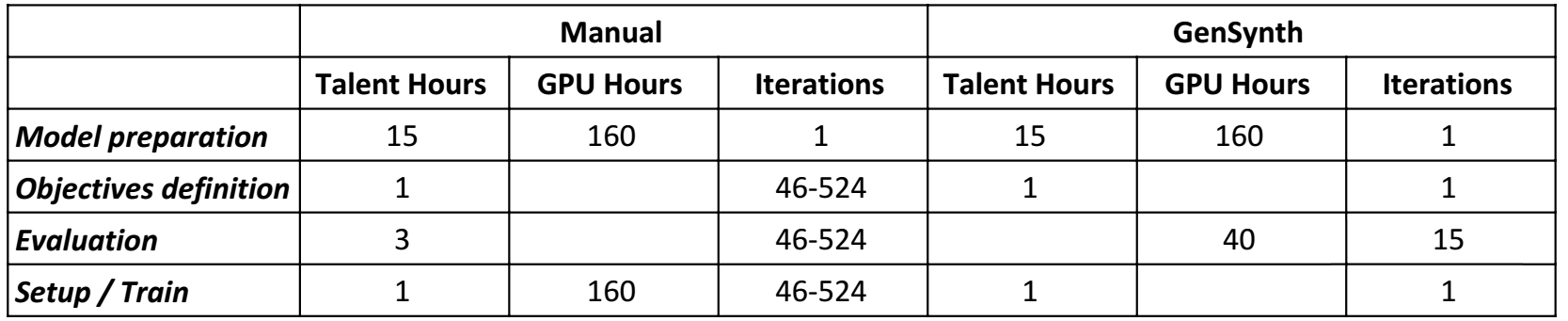}}  \\
\multicolumn{2}{c}{(a)}  \\
\includegraphics[width=6cm]{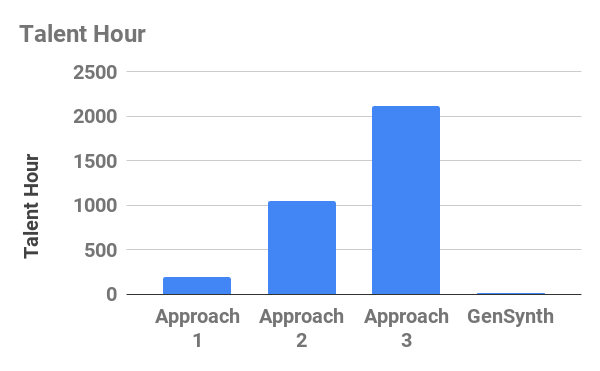}&  \includegraphics[width=6cm]{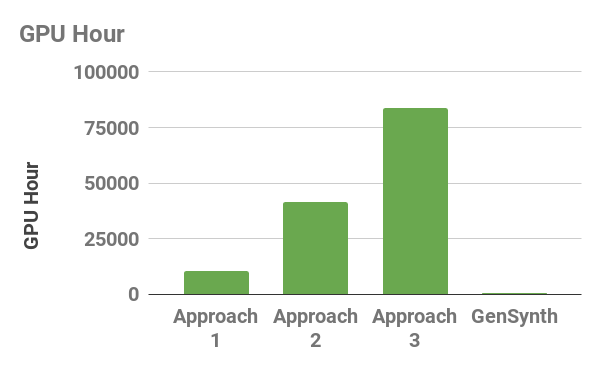}\\

((b) & (c)
\end{tabular}
\caption{\footnotesize Talent hours and GPU processing hours for manual and GenSynth approaches. (a) Shows the number of talent hours needed to prepare a  model and the  experiments that need to be performed to generate the final model for each approach. These numbers are multiplied by the number of experiments in each approach to calculate the total hours needed to create the final model shown in (b).}
\label{fig:cost_charts}
\end{figure}
 
 The number of experiments required to design the compact network depends on how many parameters need to be tuned by the user, based on the architecture of the original network. Therefore, to provide a fair and general comparison, we compare the GenSynth method with 3 different manual approaches (i.e., based on how many experiments need to be done) to design the compact network. Here we assume that each hyperparameter is evaluated twice and, to reduce the number of experiments that need to be performed, we assume that each hyperparameter is tuned independently in a sequential layer-by-layer manner:
 \begin{itemize}

  \setlength{\itemsep}{0pt plus 1pt}
 \small
     \item {\bf Approach 1:} 46  experiments based on a total of 23 hyperparameters according to different blocks in the network and considering 1 parameter per each block to tune.
     \item {\bf Approach 2:} 258 experiments based on a total number of 129 hyperparameters.
     \item {\bf Approach 3:} 524 experiments based on a total number of 262 hyperparameters.
 \end{itemize} 

For each manual experiment, 160 hours of GPU time was needed to train the network; however, GenSynth took approximately 40 hours of GPU time to complete each experiment, which includes the time needed by GenSynth to learn from past models during the inquisition phase and generate a new, fully functional, deep neural network. The GenSynth platform only needs to conduct 15 experiments to generate the final model.

 Figure~\ref{fig:cost_charts}(a) shows the estimated hours that a machine learning practitioner spends to create a model based on the training data manually compared against the GenSynth platform. While defining the objective and the model preparation are one-time tasks, Evaluation and Setup/Training are two tasks which the practitioner continually iterates  to create the final model. These two tasks take 4 hours, on average, of the practitioner's time every iteration.  However, the practitioner can offload the evaluation task to the GenSynth platform (i.e., replaced with Generate/Update/Inquire in the GenSynth platform) which saves him/her 3 hours each time. In other words, utilizing the GenSynth platform can help to conduct each experiment 4$\times$ faster. Figure~\ref{fig:cost_charts}(b) demonstrates the number of talent hours associated with each manual approach compared to the GenSynth platform.  Figure~\ref{fig:cost_charts}(c) shows the number of GPU hours each of these approaches needs to finish the task. 

 As seen in the best case scenario (i.e., the smallest number of manual experiments),  the required  talent hours for generating an optimal model is around 200 hours with 10400 GPU hours; taking advantage of the GenSynth platform would decrease the required number of talent hours to only  17 and the GPU processing hours would decrease to 760. 
 
\begin{figure}
     \centering
     \includegraphics[width=10 cm]{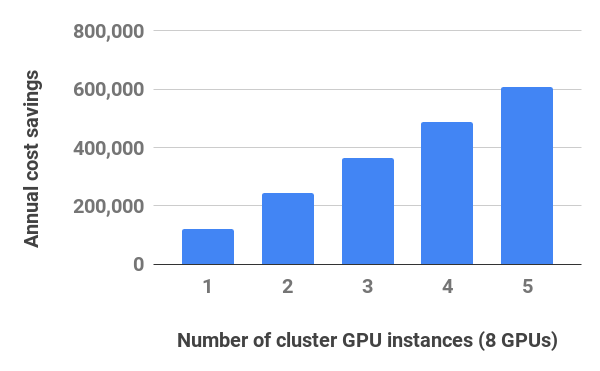}
     \caption{The annual cost savings by using the generated efficient model by GenSynth compared to the original user design prototype based on the number cluster GPU instances needed for comprehensive testing.}
     \label{fig:cost_saving}
 \end{figure}

  In addition to significantly accelerating the design of custom compact deep neural networks, as demonstrated by the above comparative analysis, a great benefit of creating highly efficient models using GenSynth is the cost savings in leveraging such models during inference time on cloud servers; this is crucial given the battery of comprehensive tests performed  to evaluate model's reliability. To better study the cost savings gained, we explore the scenario where the generated model is deployed on a cluster GPU instance (with 8 Nvidia V100 GPUs) from one of the cloud providers for comprehensive testing.  Let us assume the cost for one cluster GPU instance per hour based on on-demand policy to be 27.78\euro, and let us perform the analysis based on  the generated networks created by GenSynth platform, which was shown in experimental results to be 1.8$\times$ faster at inference when compared to the original user design prototype on a Nvidia V100 GPU, which is widely used on cluster GPU instances. As shown in Figure~\ref{fig:cost_saving}, the annual cost savings by using the generated model compared to the original user design prototype is 122,000\euro~  when only one cluster GPU instance is used. However the saved cost increases to more than $\sim$610,000\euro~  per year when 5 cluster GPU instances are used, which is typical given the battery of tests that must be performed.       

 Finally, we also assessed the usability and merits of the GenSynth platform through user feedback from AEV machine learning engineers.  A number of interesting observations were made.  First, a positive point that the engineers identified was that the web-based platform was fast and easy to use for generating optimized neural networks based on an original input design prototype, as a data scientist or engineer only had to care about which dataset to use and which training parameters they had previously used to build the input design prototype. Second, the engineers liked that the platform generates specialized neural network models which were significantly smaller and faster than the original design prototype fed into the GenSynth platform while maintaining modelling accuracy. In terms of areas of improvement, the engineers found that while the platform offers specialized training strategies, one still needs to already know the best training strategies in order to achieve the best possible outcome, thus gearing GenSynth towards machine learning experts. Moreover, since the GenSynth platform designs new neural network architectures, a practitioner still needs to validate the performance of the new network for a more thorough assessment.  Overall, the engineers at AEV had a positive experience using the GenSynth platform and were able to demonstrate the efficacy of GenSynth for accelerating compact deep neural network design.

\bibliography{egbib}
\end{document}